\documentclass[sigconf]{acmart}
\AtBeginDocument{%
  }

\copyrightyear{2025} 
\acmYear{2025} 
\setcopyright{acmlicensed}\acmConference[FAccT '25]{The 2025 ACM Conference on Fairness, Accountability, and Transparency}{June 23--26, 2025}{Athens, Greece}
\acmBooktitle{The 2025 ACM Conference on Fairness, Accountability, and Transparency (FAccT '25), June 23--26, 2025, Athens, Greece}
\acmDOI{10.1145/3715275.3732013}
\acmISBN{979-8-4007-1482-5/2025/06}

\usepackage{multirow}
\usepackage{subfigure}
\usepackage{makecell} 
\usepackage{subcaption}
\usepackage{booktabs}

\definecolor{green}{RGB}{0,146,66} 
\definecolor{blue}{RGB}{0,72,126} 
\definecolor{red}{RGB}{244,120,167}
\definecolor{orange}{RGB}{245,182,139}
\definecolor{purple}{RGB}{220,21,235} 
\definecolor{hugblue}{RGB}{0, 102, 204}

\begin{document}

\title[FairTranslate: An English-French Dataset for  Gender Bias Evaluation in MT by Overcoming Gender Binarity]{FairTranslate: An English-French Dataset for \\
Gender Bias Evaluation in Machine Translation \\
by Overcoming Gender Binarity}

\author{Fanny Jourdan}
\email{fanny.jourdan@irt-saintexupery.com}
\affiliation{%
  \institution{IRT Saint Exupery}
  \city{Toulouse}
  \country{France}
}

\author{Yannick Chevalier}
\email{yannick.chevalier@univ-lyon2.fr}
\affiliation{%
  \institution{Université Lumière Lyon 2 \\
  IHRIM UMR 5317}
  \city{Lyon}
  \country{France}}

\author{Cécile Favre}
\email{cecile.favre@univ-lyon2.fr}
\affiliation{%
  \institution{Université Lumière Lyon 2, Université Claude Bernard Lyon 1, ERIC}
  \city{69007, Lyon}
  \country{France}
}

\renewcommand{\shortauthors}{Jourdan et al.}

\begin{abstract}


Large Language Models (LLMs) are increasingly leveraged for translation tasks but often fall short when translating inclusive language -- such as texts containing the singular '\textit{they}' pronoun or otherwise reflecting fair linguistic protocols. Because these challenges span both computational and societal domains, it is imperative to critically evaluate how well LLMs handle inclusive translation with a well-founded framework.
  
This paper presents FairTranslate, a novel, fully human-annotated dataset designed to evaluate non-binary gender biases in machine translation systems from English to French. FairTranslate includes 2418 English-French sentence pairs related to occupations, annotated with rich metadata such as the stereotypical alignment of the occupation, grammatical gender indicator ambiguity, and the ground-truth gender label (male, female, or inclusive). 

We evaluate four leading LLMs (Gemma2-2B, Mistral-7B, Llama3.1-8B, Llama3.3-70B) on this dataset under different prompting procedures. Our results reveal substantial biases in gender representation across LLMs, highlighting persistent challenges in achieving equitable outcomes in machine translation. These findings underscore the need for focused strategies and interventions aimed at ensuring fair and inclusive language usage in LLM-based translation systems.

  We make the FairTranslate dataset publicly available on \textbf{{\color{blue}\href{https://huggingface.co/datasets/Fannyjrd/FairTranslate_fr}{Hugging Face}}}, and disclose the code for all experiments on \textbf{{\color{blue}\href{https://github.com/fanny-jourdan/FairTranslate}{GitHub}}}.
\end{abstract}

\ccsdesc[500]{Computing methodologies~Machine translation}

\keywords{Fairness, Natural Language Processing, Translation, LLM, Gender}


\maketitle

\section{Introduction}

\begin{figure*}
    \centering
    \includegraphics[width=1\linewidth]{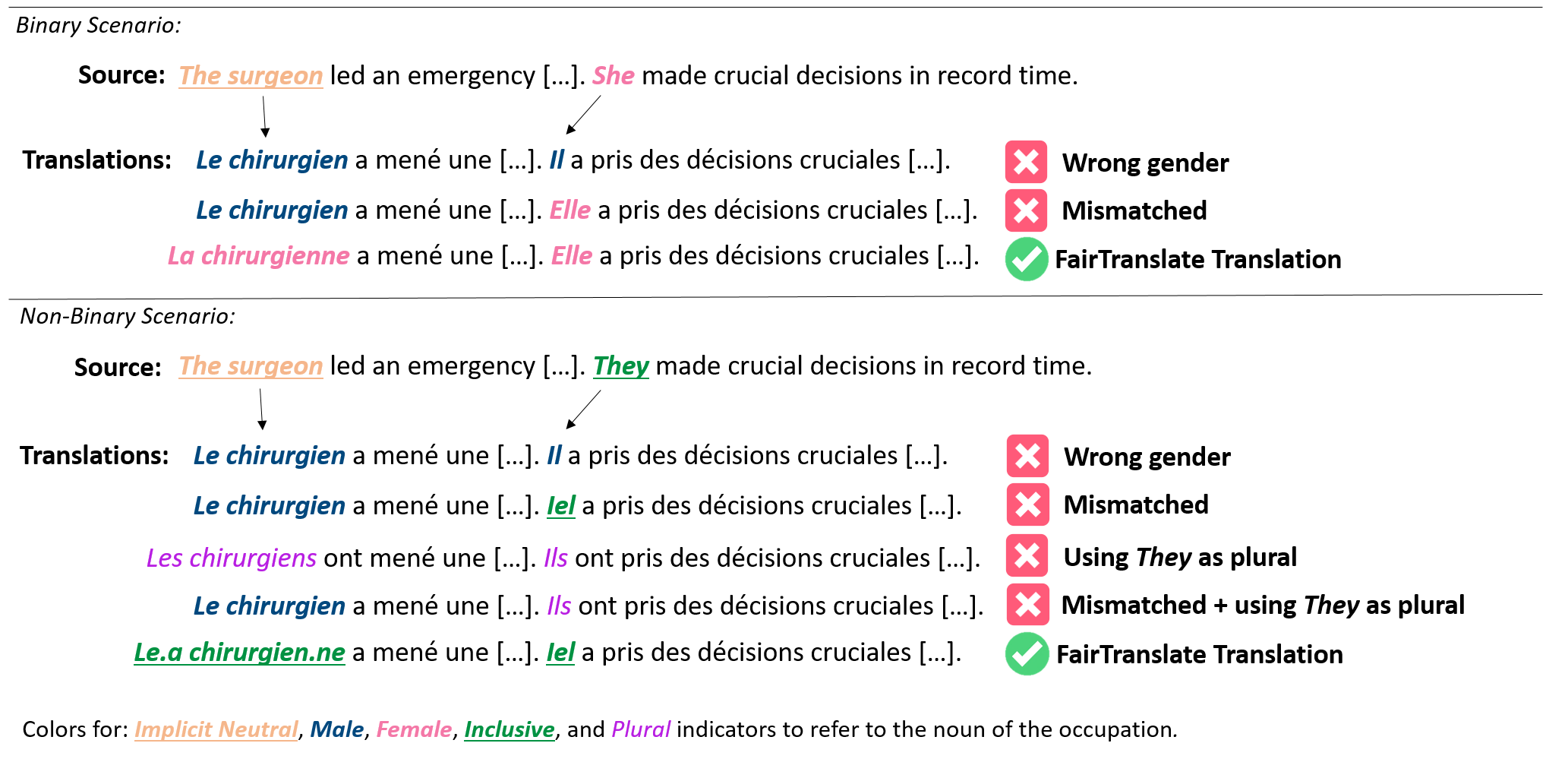}
    \Description[]{} 
    \caption{\textbf{Illustration of potential translation errors in handling gendered language in machine translation.} Starting from a source sentence in English, the binary scenario includes errors of incorrect gender assignment and mixed-gender outputs (combining correct and incorrect gender). The non-binary scenario extends these errors to include treating the singular '\textit{they}' as plural, and mixing an incorrect gender assignement with a plural. The final line shows the reference translation from the FairTranslate dataset, exemplifying an accurate treatment of inclusive language. \vspace{-4mm}}
    \label{fig:Translation_diagram}
\end{figure*}

The advent of Large Language Models (LLMs) has significantly advanced machine translation, enabling cross-linguistic communication at an unprecedented scale. However, these systems often exhibit and perpetuate social biases, particularly when handling gendered language \cite{kotek2023gender}. While much of the prior work has focused on evaluating binary gender biases, where the primary challenge is determining whether a source sentence refers to masculine or feminine forms, recent developments in inclusive linguistic practices have introduced an additional layer of complexity: accurate translation of gender-inclusive expressions.

A key translation challenge lies in the handling of the English singular '\textit{they}', which serves as an inclusive non-binary or unspecified gender pronoun. Unlike traditional binary gendered pronouns, the singular '\textit{they}' requires models to both infer that the pronoun is singular, and to preserve this gender-neutrality in their translations. Furthermore, in traditionally gendered languages such as French, articles and noun endings (for occupations, for example) are also often gendered and hence pose a challenge for fair machine translation. In addition to these two challenges, LLM translation should be able to generate gender-neutral neopronouns such as the French '\textit{iel}' (combining gender-specific '\textit{il}' (he) and '\textit{elle}' (she)). 
Figure \ref{fig:Translation_diagram} illustrates these challenges by comparing translation errors in binary and non-binary gender scenarios, highlighting the additional complexity introduced by inclusive language. 

In this work, we propose a structured approach to evaluate gender biases in machine translation. We develop and publicly share FairTranslate, a novel English-French translation dataset designed for assessing both binary and non-binary gender biases. 
The dataset is composed of sentences related to specific occupations. It is richly annotated with metadata, including gender labels (male, female, or inclusive), markers of gender ambiguity (in the english sentence), and indicators of alignment with stereotypical occupational gender-roles. This detailed annotation enables fine-grained analyses of model performance, particularly in their treatment of inclusive language.

Using FairTranslate, we conduct an evaluation of four widely used LLMs (Gemma2 \cite{team2024gemma}, Mistral \cite{jiang2023mistral}, Llama3.1, and Llama3.3 \cite{dubey2024llama}) across four prompting strategies, including moral and linguistic promptings. Our findings reveal that current machine translation systems struggle significantly with inclusive language. Translation quality for inclusive forms consistently lags behind that of binary forms. We observe that French inclusive markers such as '\textit{iel}' are almost never generated, even with tailored prompting. Importantly, we show that these failures are not solely due to the recent introduction of inclusive practices into the French language. Instead, they reflect deeper issues in the ability of LLMs to represent, interpret and translate established inclusive constructs in English, such as the singular '\textit{they}'.



\section{Background}\label{sec:background}

From a French linguistic point of view, grammatical gender is a tool that operates according to two main principles \cite{corbett1991gender}:

\textit{Principle A}: Grammatical gender ensures the internal cohesion of sentences. A donor element (the noun, which has a fixed gender) imposes gender agreement on satellite elements (adjectives, determiners, pronouns) \cite{chevalier2017bases}. 

\textit{Principle B}: In the case of nouns referring to human beings (such as in occupation titles), grammatical gender helps convey the gender identity of the person who is being discussed. 

Depending on the language, these two main principles may be applied differently. This is particularly evident in the comparison between French (a language with highly pronounced grammatical gender) and English (a language where grammatical gender is considered less prominent) \cite{chevalier2017bases}. The challenges of developing non-discriminatory writing protocols are therefore different in the two languages. While the non-discriminatory protocols of English are relatively stable and grammaticalized, they are less established in French where grammatical gender is more complex \cite{perez2019des, abbou2013pratiques}. The challenges which are introduced by such differences in grammatical gender across languages are significant and must be explicitly addressed. Specifically:

(i) In English, \textit{principle A} does not apply. The gender of nouns referring to human beings does not affect the satellite elements that refer to the noun (e.g., determiners: '\textit{a nurse, the nurse}'; adjectives: '\textit{the professional nurse}'; or verb forms: '\textit{the nurse has come}'). In contrast, in French, the grammatical gender of the noun influences satellite elements such as determiners (\textit{'un'/'une', 'le'/'la'}), adjectives (\textit{'professionnel'/'professionnelle'}), and compound verb forms (\textit{'il est venu'/'elle est venue'}) \cite{elmiger2017binarite}.

(ii) In English, \textit{principle B} rarely applies as most occupations names tend to be gender-invariant. In fact, the gender of nouns referring to human beings is rarely marked in English (e.g., '\textit{nurse}'). However, in French, gender of nouns referring to human being is often marked through alternating suffixes (e.g., \textit{'infirmier'} meaning male nurse, and \textit{'infirmière'} meaning female nurse) \cite{abeille2021grande}.

(iii) Gender in pronouns is marked differently in English and French. In English, third-person singular pronouns (\textit{'he'/'she'}) indicate gender, but the third-person plural pronoun (\textit{'they'}) does not. In contrast, French marks gender in both third-person singular (\textit{'il'/'elle'}) and plural (\textit{'ils'/'elles'}) forms. Additionally, English possessive determiners inherit the gender of their antecedent (e.g., '\textit{his/her dog}' from '\textit{he/she owns a dog}'), whereas French possessives agree with the gender of the noun they determine (e.g., '\textit{son chien}' for a male dog and '\textit{sa chienne}' for a female dog, regardless of the owner's gender).

(iv) Finally, there is a particular difficulty in English in distinguishing between the two main uses of \textit{they}: first as a third-person plural pronoun (\textit{Plural They}) and second as an inclusive third-person singular pronoun (\textit{Singular They} or \textit{Inclusive They}). This issue does not arise in French, where separate pronouns are available for third-person singular ('\textit{iel}' or equivalents) and third-person plural ('\textit{iels}' or equivalents) to avoid expressing grammatical gender.

Machine translation from English to French must therefore propose satisfactory solutions to the following challenges: 

(i') Once the correct French target word has been identified, the machine translation system must apply gender agreement rules to all related satellite elements (determiners, adjectives, verb forms), as well as to the pronouns used.

(ii') For English source terms (e.g., \textit{'nurse'}), machine translation systems should generate three corresponding forms in French: the masculine form (\textit{'infirmier'}), the feminine form (\textit{'infirmière'}), and the inclusive form (\textit{'infirmier.ière'} or an equivalent). This selection can be guided by contextual cues in the English text, such as pronouns (\textit{'he'/'she'/inclusive 'they'}) and possessive determiners (\textit{'his'/'her'/inclusive 'their'}). This process is analogous to performing coreference resolution: the task of identifying all expressions in a text that refer to the same entity, often focusing on how pronouns are linked to their antecedents.
However, even when such information is available—and especially when no contextual cues are present to guide the model—the system may fail to appropriately gender the target term. This issue partly stems from the historical perception of the masculine form as "neutral" in the French linguistic tradition, aligned with the conservative recommendations of the \textit{Académie Française}. Nonetheless, the widespread claim that feminine forms do not always exist is inaccurate: the morphology of French is highly flexible in forming feminine forms, as evidenced by their presence in the earliest written texts in the language\footnote{See a list of all historically attested feminine forms in French at \url{https://siefar.org/la-guerre-des-mots/presentation/}}.
The tendency of models to predominantly use masculine forms (or feminine forms for certain highly stereotyped professions) reinforces existing gendered representations \cite{gygax12masc}. It is crucial to prevent quantitative descriptions—which are themselves products of social constructions—from acting as descriptive norms that evolve into prescriptive norms. This phenomenon is particularly significant in contexts such as influencing young people's career choices, where such biases may perpetuate gender stereotypes \cite{gygax2021cerveau}.

(iii') and (iv') The differences in pronoun usage between English and French require machine translation systems to distinguish between singular '\textit{they}' (person 3) and plural '\textit{they}' (person 6). This distinction involves accurately identifying whether the antecedent in the English source text is singular or plural. Once this has been determined, translating into French is generally straightforward for singular '\textit{they}' (\textit{Inclusive They} = '\textit{iel}'). However, difficulties persist for plural 'they,' as the system must select among the three available forms in French (\textit{'ils'/'elles'/'iels'}) based on the desired level of inclusivity.

\section{Related Work}
Evaluating gender bias in Machine Translation (MT) has been extensively studied using benchmarks like Winogender \cite{rudinger2018gender}, Winobias \cite{zhao2018gender}, WinoMT \cite{stanovsky2019evaluating} and, more recently, WinoPron \cite{gautam-etal-2024-winopron}, which assess coreference resolution to determine how models attribute gender based on context. Similarly, MT-GenEval \cite{currey2022mt} employs a counterfactual methodology by systematically modifying sentences to analyze gender biases. Our approach integrates both coreference resolution and counterfactual evaluation, extending these methodologies beyond the binary framework to include non-binary gender and address the broader challenges associated with it.

Recent studies addressing non-binary gender \cite{felkner2023winoqueer, dhingra2023queer, waldis-etal-2024-lou} focus on evaluating bias related to gender and LGBTQ+ identities, but do not specifically target machine translation. AmbGIMT \cite{chen2024beyond} evaluates non-binary gender in English-Chinese MT, but its focus is on attitude translation rather than coreference or the use of singular they. An English-German MT study also exists, focusing on gender-neutral person-referring terms \cite{lardelli-etal-2024-building}.

While these works cover English-Chinese and English-German, the addition of an English-French dataset enables exploration of non-binary gender translation across typologically diverse languages such as French, German, and Chinese, opening up new research opportunities.

\section{FairTranslate Dataset}

Building on the considerations outlined in earlier sections, the FairTranslate Dataset includes \textbf{2,418 entries} designed to investigate how LLMs handle gender in English-to-French translation, paying special attention to inclusive forms and subtler expressions of gender bias. 

FairTranslate is composed of English-French sentence pairs, each centered around an occupation to investigate how gender is expressed and translated. Each sentence is annotated with a gender label (male, female, or inclusive) corresponding to the individual referenced by the occupation. In addition, the dataset includes rich metadata, detailed in the following section.

\begin{figure*}
    \centering
    \includegraphics[width=1\linewidth]{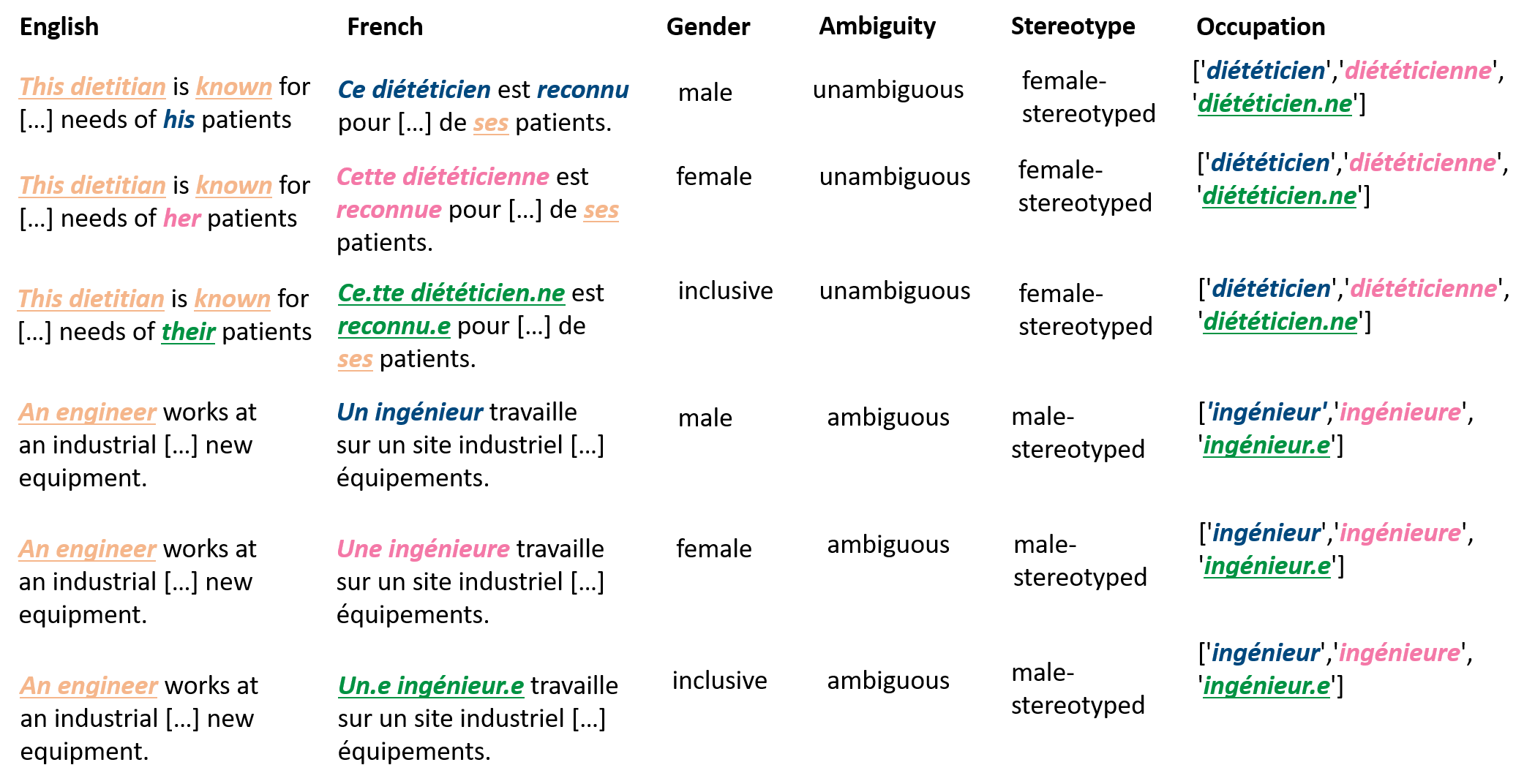}
    \caption{Examples from the FairTranslate dataset. Words in colors indicate \textcolor{orange}{\underline{\textbf{\textit{ambiguous/neutral}}}}, \textcolor{blue}{\textbf{\textit{male}}}, \textcolor{red}{\textbf{\textit{female}}}, and \textcolor{green}{\underline{\textbf{\textit{inclusive}}}} entities.}
    \label{fig:dataset_example}
\end{figure*}

To enable counterfactual interventions and direct comparisons, each example appears in all three gender variants (male, female, and inclusive). This design facilitates the evaluation of gender-specific translations and supports research on coreference resolution. Examples of sentences from the dataset, along with their corresponding annotations, are illustrated in Figure \ref{fig:dataset_example}.

In combination, these design choices produce a dataset that can serve as a robust benchmark for evaluating fairness, inclusivity, and nuanced coreference resolution in modern translation models.
The result is a richer, more controlled environment for studying fairness and inclusivity in machine translation.

\subsection{Structure}

Each entry in the FairTranslate dataset corresponds to a pair of English-French sentence translations, linked to an occupation, and annotated with several attributes. These attributes reflect the key variables in our study, enabling counterfactual comparisons and nuanced analyses of model behavior.
\begin{itemize}
    \item \textbf{English}:
    English sentences which involve an occupation. These sentences were chosen to examine how models respond to both entrenched stereotypes as well as to more nuanced contexts.
    \item \textbf{French}:
    French translations, which are faithful to the gender of the English original. These sentences serve as a ground-truth reference for the English-French translation task.
    \item \textbf{Gender (male / female / inclusive)}:
    This column indicates the intended grammatical gender of the occupation noun in the French translation. 
    \item \textbf{Ambiguity (ambiguous / unambiguous / long unambiguous)}:
    This column specifies the clarity of the referent’s gender in the English original.
    \begin{itemize}
        \item \textit{ambiguous}: No explicit cues in English indicate the referent’s gender, allowing multiple valid interpretations (masculine, feminine, or inclusive) in French.
        \item \textit{unambiguous}: Pronouns or other linguistic cues make the referent’s gender immediately clear (e.g., '\textit{he}', '\textit{she}' or '\textit{they}' in close proximity to the target noun).
        \item \textit{long unambiguous}: The gender can be inferred, but only from contextual cues appearing after a delay (e.g., multiple sentences later), thus testing a model’s ability to perform coreference resolution over longer stretches of text.
    \end{itemize}
    By differentiating these categories, we can analyze whether models default to masculine forms in ambiguous cases and whether they can accurately resolve pronouns when explicit signals are present, either nearby or further afield.
    \item \textbf{Stereotype (male-stereotyped / female-stereotyped / gender-balanced)}:
    Each sentence refers to an occupation assigned to one of these categories, based on real-world gender distribution data\footnote{The gender distribution data for occupation was sourced from Statbel, the official Belgian statistical office.}. Professions statistically dominated by men (e.g., '\textit{mechanic}'), dominated  by women (e.g., '\textit{nurse}') or gender-balanced (e.g., '\textit{attorney}') challenge models to translate consistently regardless of stereotype. 
     \item \textbf{Occupation (list of gendered forms in French)}: For each occupation noun, this column contains a list of its three gendered forms in French (male, female, and inclusive). For example, if the English original refers to a '\textit{nurse}', the corresponding value is ['\textit{infirmier}', '\textit{infirmière}', '\textit{infirmier.ière}']. This information enables precise evaluation of models' ability to accomplish grammatical gender agreement across different linguistic forms while maintaining the the intended meaning. 
\end{itemize}


\subsection{Dataset Construction}

\begin{figure}[ht]
\centering

\textbf{Feminine}\\
\texttt{[“dietician”, “cleaner”, “schoolteacher”, “childminder”, “nursing assistant”, “nurse”, “pharmacy assistant”, “hairstylist”, “beautician”, “cashier”, “teller”, “accounting employee”, “social worker”, “pharmacist”, “salesperson”, “flight attendant”, “childcare worker”, “caregiver”, “daycare worker”, “housekeeper”, “secretary”, “librarian”]}\\[0.5em]

\textbf{Masculine}\\
\texttt{[“construction worker”, “logger”, “firefighter”, “electrician”, “welder”,  “plumber”, “mechanic”, “carpenter”, “joiner”, “electromechanic”, “street sweeper”, “garbage collector”, “butcher”,  “engineer”, “bus driver”, “supervisor”, “computer scientist”, “programmer”, “police officer”, “surgeon”, “construction machine operator”, “upholsterer”]}\\[0.5em]

\textbf{Neutral}\\
\texttt{[“chemical technician”, “management controller”, “press operator”, “buyer”, “quality technician”, ”physiotherapist”, “lawyer”, “server”, “teacher”, “product manager”, “translator”, “project manager”, “career counselor”, “doctor”, “optician”, “special education teacher”, “journalist”, “accountant”]}\\

\caption{Lists of occupations used for sentence generation, grouped by stereotypical gender.}
\label{fig:occupation_lists}
\end{figure}

We adopted a two-step approach to generate and annotate the FairTranslate Dataset. First, we designed target sentences in French. This has enabled us to capture all the relevant gender-marking structures inherent to French. Second, we translated these French sentences into English while maintaining consistency across the gender variants.

\textbf{Step 1: French Sentence Generation and Annotation}

\begin{itemize}
    \item Selecting Occupations (see Figure \ref{fig:occupation_lists}):
We begin with three curated lists of occupations: \begin{itemize}
    \item Male-stereotyped (22 occupations predominantly held by men)
    \item Female-stereotyped (22 occupations predominantly held by women)
    \item Gender-balanced (18 occupations held by men and women at similar rates)
\end{itemize}
This selection served a dual purpose: it enabled the construction \textbf{stereotype} and \textbf{occupation} variables, and it produced a diverse set of sentences for testing machine translation across a range of occupational contexts. 

    \item Sentence Creation with GPT Assistance:
Using GPT-4o \cite{openai2024gpt4o} and GPT-o1 \cite{openai2024o1} under close human supervision, we generated French sentences that captured each occupation and related activities. A human operator prompted and reviewed the model outputs, making manual edits as needed to ensure variety, relevance, and proper structure. This hands-on oversight was critical to guarantee high-quality and contextual validity of the sentences.

    \item Ambiguity Annotations for English Sentences: For each occupation, we generated three types of sentences: \begin{itemize}
        \item Ambiguous (5 sentences): No explicit pronoun or other cue indicates the person’s gender for English translations. French sentences were by nature unambiguous since the form of the occupation is gendered.
        \item Unambiguous (5 sentences): A pronoun or nearby context reveals the gender for the English translation.
        \item Long Unambiguous (3 sentences): Gender is determinable only after multiple lines or sentences of text, testing longer-range coreference resolution for the English translation.
    \end{itemize}
    These categories allowed us to label each example with the “ambiguity” variable, ensuring clarity regarding when and how gender clues appear in the text.
    
    \item Gender Variants in French: Each French sentence was replicated in masculine, feminine, and inclusive forms. We did this by making changes only to the occupational forms and the related satellite elements, while keeping other elements constant across variants. The \textbf{Occupation} column was used to systematically retrieve and apply the correct gendered forms for each occupation. This method ensured that the sentences remained semantically identical across the three gender variants. 
    
    \item Final Checks and Annotation: Human annotators verified that each sentence was correctly annotated across the variables. This process has helped ensure the integrity and consistency of the data prior to translation.
\end{itemize}

\textbf{Step 2: English Translation}

Once the French sentences and their annotations (gender, ambiguity, stereotype, occupation) were finalized, we used GPT-4o to translate them into English, adhering to the following guidelines: \begin{itemize}
    \item Maintaining Consistency Across Gender Variants: For each set of three French sentences (masculine, feminine, inclusive) belonging to the same base example, we produced aligned English translations in which all aspects remained identical except for the explicitly gendered elements (e.g., '\textit{he}', '\textit{she}', or singular '\textit{they}' pronouns; possessive forms '\textit{his}', '\textit{her}', '\textit{their}')\footnote{This only applies to possessive forms in English, as in French, possessive forms do not agree with the possessor.}.
    
    \item Human Verification: We tasked a human reviewer with the quality assessment of the automated translations. In particular, the task was to confirm the accuracy of the translation, and ensure that no changes beyond the gender markers had been introduced. This step has helped ensure consistent pairing between the French sentences and their English counterparts.
\end{itemize}


\subsection{The Challenges of Inclusive Language}

Inclusive language refers to ways of expressing oneself that aim to ensure women are not excluded in occupational contexts, individuals are addressed appropriately either with respect to or regardless of their gender identity (male, female, or non-binary), and gender diversity is highlighted when referring to mixed groups. These aims adapt to the linguistic characteristics of each language.

In English, proponents of inclusive language have proposed a variety of forms over the years, including neopronouns such as \textit{'xe'} or \textit{'ze'}. However, the singular \textit{'they'}, along with the possessive \textit{'their'} and gender-neutral occupational terms (e.g., \textit{'Chair'} instead of \textit{'Chairman'} or \textit{'Chairwoman'}), has emerged as the most widely accepted standard due to its historical roots and frequent contemporary use. For this reasons, we have opted to use this standard in the construction of our dataset.

In French, inclusive language is far less standardized, where the diverse existing practices have been described as a “graphic tumult” \cite{abbou2013pratiques}. 
For occupational nouns, French speakers tend to use lexical doublets (e.g. '\textit{enseignant}' (male) and '\textit{enseignante}' (female)). Inclusive forms, on the other hand, are represented with a wide variety of typographical signs (\textit{'enseignant.e', 'enseignant/e', 'enseignant-e', 'enseignantE'}). Similarly, inclusive neopronouns (\textit{'iel'/'al'/'ul'}) and determiners (\textit{'lea'/'la·le'/'lo'}) are still subject to great variability.  Efforts to systematize these practices include works like \citet{2018grammaire, alpheratz:hal-02323626, touraille2023chapitre}, Hadad's guidelines\footnote{\url{https://www.motscles.net/ecriture-inclusive}}, and the recommendations of the HCEHF\footnote{\url{https://www.haut-conseil-egalite.gouv.fr/IMG/pdf/guide_egacom_sans_stereotypes-2022-versionpublique-min-2.pdf}}. Other innovative approaches include inclusive typography, as explored by \citet{circlude2023typographie} and the ByeByeBinary collective’s type library\footnote{\url{https://typotheque.genderfluid.space/fr}}.

In the construction of the FairTranslate dataset, we opted for the following inclusive French writing model. We used a selected form for each occupational term and its related elements. For the singular \textit{'they'}, we opted for French \textit{'iel'}, paring the indefinite English article \textit{'a(n)'} with French \textit{'un.e'}, and the definite \textit{'the'} article with French \textit{'lea'}. 
Due to the lack of standards, we had to make choices in the construction of FairTranslate. In order to accommodate for other French inclusive writing practices, we developed a Python dictionary made available on GitHub\footnote{\url{https://github.com/fanny-jourdan/FairTranslate}}. The dictionary provides an easy way to map our chosen inclusive forms to a wider range of recognized alternatives. The dictionary enables machine translations to be evaluated as correct even if they use a different inclusive form than those explicitly listed in the dataset (e.g., \textit{'ul'} instead of \textit{'iel'}). By offering this adaptable tool, we aim to support inclusive translation research while remaining receptive to the natural evolution of inclusive language.
 
\begin{table*}[ht]
    \centering
    \begin{tabular}{lcccccccccc}
        \toprule
        \multirow{2}{*}{Model} & \multicolumn{2}{c}{Gemma2-2B} & \multicolumn{2}{c}{Mistral-7B} & \multicolumn{2}{c}{Llama3.1-8B} & \multicolumn{2}{c}{Llama3.3-70B} & \multicolumn{2}{c}{Mean} \\
        \cmidrule(lr){2-3} \cmidrule(lr){4-5} \cmidrule(lr){6-7} \cmidrule(lr){8-9} \cmidrule(lr){10-11}
         & BLEU & COMET & BLEU & COMET & BLEU & COMET & BLEU & COMET & BLEU & COMET \\
        \midrule
        Female & 42.34 & 87.86 & 38.94 & 87.16 & 43.73 & 88.88 & 51.44 & 90.12 & 44.16 & 88.56 \\
        Male & \textbf{45.37} & \textbf{89.18} & \textbf{41.36} & \textbf{88.25} & \textbf{46.18} & \textbf{89.85} & \textbf{55.86} & \textbf{91.35} & \textbf{47.19} & \textbf{89.66} \\
        Inclusive & 36.65 & 85.42 & 33.77 & 84.42 & 37.53 & 85.97 & 45.27 & 87.16 & 38.35 & 85.80 \\
        \bottomrule
    \end{tabular}
    \caption{Comparison of BLEU and COMET scores (in percentages) for different genders and models, with mean scores across models for Baseline Prompting. The bold font indicates the gender for which the model performs best. We observe that models show best performance on the male gender, and worst performance on the inclusive gender. This finding does not vary across models or translation quality metrics. \vspace{-4mm}}
\label{tab:gender_model_comparison_with_means}
\end{table*}

\section{Experimental Setup}

As introduced in the previous section, we use the FairTranslate dataset specifically designed to evaluate gender bias in translation. The goal of this section is not to fine-tune or train the models, but to use them in a zero-shot evaluation setting. We pass all English sentences from FairTranslate to each of the selected LLMs and collect their French translations. This allows us to systematically assess how each model handles gendered and inclusive language without any additional training.

For each experiment, we evaluate a range of open-source LLMs with varying sizes and architectures to ensure broad coverage and generalizability of our findings. The selected models include: Gemma2 2B \citep{team2024gemma}; Mistral 7B \citep{jiang2023mistral}; Llama3.1 8B, and Llama3.3 70B (equivalent to Llama3.1 405B) \citep{dubey2024llama}.

We introduce several prompting strategies:

\begin{itemize}
\item \textbf{Task prompting}: We define task prompting as the baseline instruction for the translation task:
\end{itemize}
\begin{quote} ``\textit{Translate the following sentences from English to French:} '\{english\_sentence\}'. \textit{Respond with the translation only, nothing else.}'' \end{quote}

The task prompting serves as a simple directive for the model to perform the translation without any additional guidance or constraints. It will be used as a \textbf{Baseline} in our experiments. Building on this, we introduced three prompting strategies to encourage gender-inclusive translations.

We hypothesized that the observed gender biases in the generated translations can be attributed to a lack of moral and linguistic awareness, as previously suggested in studies such as \citet{hansen2022social} and \citet{zhao2021ethical}. To account for these possibilities, we introduced additional prompting strategies designed to elicit moral and linguistic knowledge:

\begin{itemize}
\item \textbf{Moral prompting}: To promote gender-inclusive and inclusive translations, we provide the following instruction --proposed by \citet{chen2024beyond}-- before the task prompting:
\begin{quote}
``\textit{You are a translation model without gender bias and LGBTQA+ friendly.}'' + task prompting.
\end{quote}

\item \textbf{Linguistic prompting}: To explicitly encourage the use of gender-inclusive linguistic forms in French, we append the following instruction:
\begin{quote}
    ``\textit{Forms like 'iel' as a neutral pronoun, 'un·e,' 'lea,' or 'ce·tte' as neutral determiners, or a mid-dot (e.g., 'étudiant·e') for gender-neutral terms to be applied only if explicitly requested.}'' \footnote{In addition to the context linguisitc, we were obliged to add this sentence: "\textit{Otherwise, use the classic feminine or masculine form.}" to counter the LLMs' bias of putting all sentences in neutral because the prompt was talking about it.} + task prompting.
\end{quote}

\item \textbf{Moral and Linguistic prompting}: This strategy combines the moral and linguistic promptings to provide both ethical and grammatical guidance for gender-inclusive translations:
\begin{quote}
Moral prompting + linguistic prompting + task prompting.
\end{quote}
\end{itemize}

By evaluating translations under these three promptings --- \textit{moral prompting}, \textit{linguistic prompting}, and \textit{moral and linguistic prompting} --- in addition to the baseline task prompting, we aim to quantify their respective impacts on reducing gender bias.

\section{General Translation Performance}
\label{sec:translation_evaluation}

This section evaluates the general translation performance of four leading Large Language Models (LLMs) on the FairTranslate dataset. Our primary objective is to assess the translation quality across gender categories (male, female, and inclusive) using two complementary metrics: BLEU and COMET. These metrics provide a dual perspective on translation accuracy, balancing surface-level quality and deeper semantic correctness. By analyzing the results, we aim to identify patterns in how LLMs handle translations involving gendered and inclusive language, as well as disparities between gender categories.

\subsection{Translation Evaluation Metrics}
We adopt two widely recognized metrics to evaluate translation quality. \textbf{BLEU} \citep{papineni2002bleu} measures n-gram overlap between machine-generated and reference translations. While it provides a quick and interpretable measure of translation quality, it is limited in capturing semantic nuances. In contrast, \textbf{COMET} \citep{rei-etal-2020-comet}, specifically the wmt22-comet-da model \citep{rei2022comet}, is a neural-based metric fine-tuned on human-annotated datasets to assess semantic alignment and translation quality. COMET captures subtle linguistic variations, including gender-related adaptations, and demonstrates stronger correlations with human judgment compared to BLEU. The use of both metrics enables a balanced evaluation and allows for the analysis of translation performance from multiple dimensions.



\subsection{Global Results}
Table \ref{tab:gender_model_comparison_with_means} reports BLEU and COMET scores for male, female, and inclusive gender categories across the four LLMs. Across all models, scores are highest for the male gender, followed by the female gender, and lowest for the inclusive gender. ANOVA tests (Table \ref{apx:tab:anovatest}) indicate that these differences are statistically significant, highlighting a systemic disparity in translation quality. As each dataset example exists in all three gender variants, the observed differences can be attributed to the model performance rather than to the dataset bias. These findings emphasize persistent challenges in achieving gender-inclusive translation, with LLMs consistently underperforming on translation of gender inclusive forms.


\subsection{Effects of Prompting}

\begin{table*}[h!]
\centering
\begin{tabular}{lcccccccccccc}
\toprule
\textbf{Prompting} & \multicolumn{3}{c}{\textbf{Gemma2-2B}} & \multicolumn{3}{c}{\textbf{Mistral-7B}} & \multicolumn{3}{c}{\textbf{Llama3.1-8B}} & \multicolumn{3}{c}{\textbf{Llama3.3-70B}} \\
\cmidrule(lr){2-4} \cmidrule(lr){5-7} \cmidrule(lr){8-10} \cmidrule(lr){11-13}
 & Female & Male & Inclusive 
 & Fem. & Male & Incl. & Fem. & Male & Incl.& Fem. & Male & Incl. \\
\midrule
Baseline & \textbf{42.34} & \textbf{45.37} & 36.65 & \textbf{38.94} & \textbf{41.36} & \textbf{33.77} & \textbf{43.73} & \textbf{46.18} & 37.53 & \textbf{51.44} & \textbf{55.86} & 45.27 \\
Moral & 42.04 & 45.03 & 37.33 & 37.43 & 40.62 & 33.12 & 42.03 & 45.49 & 37.76 & 45.02 & 50.40 & 43.87 \\
Linguistic & 42.11 & 44.39 & \textbf{37.70} & 36.78 & 40.81 & 33.48 & 42.96 & 44.68 & 36.98 & 51.40 & 55.32 & 45.28 \\
Moral \& Ling. & 42.11 & 44.70 & 37.08 & 36.36 & 39.29 & 32.22 & 42.82 & 45.24 & \textbf{38.26} & 50.57 & 55.27 & \textbf{45.94}  \\
\bottomrule
\end{tabular}
\caption{BLEU scores for different promptings and genders. The bold font indicates the prompting strategy which performs best for each model-gender combination. \vspace{-4mm}}
\label{tab:prompting_bleu_scores}
\end{table*}

Table \ref{tab:prompting_bleu_scores} evaluates the translation performance of four prompting strategies (baseline, moral, linguistic, and moral+linguistic) as indicated by BLEU scores across gender categories. Results using COMET scores are provided in the Appendix \ref{apx:sec:prompting_comet}.

For the male and female genders, translation performance generally declines as more information is incorporated into the prompts. For the inclusive gender, we observed performance decreases for Mistral and performance improvements for Gemma2, Llama3.1, and Llama3.3 models. Overall, translation performance tends to decrease for the traditional binary genders (male and female) and increase for the inclusive gender as the prompting strategies become more informative. These findings largely reflect the fact that our prompting strategies were designed the improve the translation performance of inclusive gender forms. The same prompts which are successful at that task appear to create noise for the translation of traditional gender forms. Notably, while the performance on both female and male categories declined with prompting, the decline was steeper for male gender. This had the effect offurther narrowing the performance gap which was initially observed between the male and female gender translations. Consequently, the overall disparity among all gender categories is reduced. In spite of this reduction, the male category still had better translation results, at statistically significant rates. This indicates that complete equity in machine translation has not yet been achieved.

\subsection{Analysis of Metadata Variables}
\label{sec:metavar}

\begin{table}[ht]
\centering
\begin{tabular}{|llcc|}
    \toprule
    \multirow{2}{*}{\parbox{1.7cm}{\centering \makecell{Stereotype}}} & \multirow{1.6}{*}{Gender} & \multicolumn{2}{c|}{Gemma2-2B} \\
    \cmidrule(lr){3-4} 
     &  & BLEU & COMET \\
    \midrule
    \multirow{4}{*}{\makecell{Female-\\stereotyped}} & Female & 40.78 & 87.84 \\
     & Male & \textbf{42.35} & \textbf{88.33} \\
     & Inclusive & 34.88 & 84.60 \\
     \cmidrule(lr){2-4}
     & \textit{Mean} & \textit{39.34} & \textit{86.92} \\
    \midrule
    \multirow{4}{*}{\makecell{Male-\\stereotyped}} & Female & 41.33 & 87.13 \\
     & Male & \textbf{44.69} & \textbf{88.76} \\
     & Inclusive & 36.45 & 85.42 \\
     \cmidrule(lr){2-4}
     & \textit{Mean} & \textit{40.82} & \textit{87.10} \\
    \midrule
    \multirow{4}{*}{\makecell{Gender-\\balanced}} & Female & 45.29 & 88.72 \\
     & Male & \textbf{49.63} & \textbf{90.63} \\
     & Inclusive & 38.91 & 86.36 \\
     \cmidrule(lr){2-4}
     & \textit{Mean} & \textbf{\textit{44.61}} & \textbf{\textit{88.57}} \\
    \bottomrule
\end{tabular}
\vspace{0.5em} 
\begin{tabular}{|llcc|}
    \toprule
    \multirow{2}{*}{\parbox{1.5cm}{\centering \makecell{Ambiguity}}} & \multirow{2}{*}{Gender} & \multicolumn{2}{c|}{Gemma2-2B} \\
    \cmidrule(lr){3-4} 
     &  & BLEU & COMET  \\
    \midrule
    \multirow{4}{*}{Ambiguous} & Female & 36.65 & 88.25  \\
     & Male & \textbf{41.32} & \textbf{89.36}  \\
     & Inclusive & 35.17 & 85.78 \\
     \cmidrule(lr){2-4}
     & \textit{Mean} & \textit{37.71} & \textbf{\textit{87.80}}  \\
    \midrule
    \multirow{4}{*}{\makecell{Long\\Unambiguous}} & Female & 50.35 & 87.66 \\
     & Male & \textbf{51.66} & \textbf{88.77}  \\
     & Inclusive & 45.48 & 86.67  \\
     \cmidrule(lr){2-4}
     & \textit{Mean} & \textbf{\textit{49.16}} & \textit{87.69} \\
    \midrule
    \multirow{4}{*}{Unambiguous} & Female & 43.17 & 87.60  \\
     & Male & \textbf{45.61} & \textbf{89.25}  \\
     & Inclusive & 32.76 & 84.30  \\
     \cmidrule(lr){2-4}
     & \textit{Mean} & \textit{40.51} & \textit{87.05}  \\
    \bottomrule
\end{tabular}
\caption{\textbf{BLEU and COMET scores for the Gemma2-2B model and each combination of (Top) stereotype and (Bottom) ambiguity with gender.} Results are obtained with Baseline Prompting. Bold font highlights the best-performing gender per group. Bold italicized font shows the best-performing stereotype or ambiguity category. Additional results appear in Appendix \ref{apx:sec:metavar}.}
\label{tab:stereo_ambig_scores_gemma2}
\end{table}

To analyze biases in more detail, we examine the effects of occupational stereotypes and sentence ambiguity on translation performance for Gemma2-2B. 

Table \ref{tab:stereo_ambig_scores_gemma2} shows the performance of Gemma2 across gender categories (male, female, inclusive) and two key variables: \textit{Stereotype} and \textit{Ambiguity}. The left facet reports BLEU and COMET scores for sentences categorized by occupational stereotypes (female-stereotyped, male-stereotyped, gender-balanced). The right facet provides a breakdown of the ambiguity categories (ambiguous, long unambiguous, unambiguous). Results for other models, provided in Appendix \ref{apx:sec:metavar}, demonstrate consistent behavior across these variables. Across all stereotype and ambiguity categories, the instances with male gender were consistently better translated than those with other genders, irrespective of the variable examined.

For stereotype categories, even for female-stereotyped occupations, male translations outperformed female translations. While the gap between male and female scores was the smallest in this category, less than 2\% in BLEU and less than 1\% in COMET, it is notable that female gender translations did not achieve higher scores, as one might expect given the stereotypical association. This could be due to the historical tendency in French to default to the masculine form for occupations, even when those occupations are predominantly associated with women. 
Unsurprisingly, male-stereotyped and gender-balanced occupations showed a larger gap. In all cases, scores for the inclusive gender were lower than those for male and female genders, with the largest differences observed in gender-balanced occupations (11\% in BLEU and 4\% in COMET). The generally superior performance in gender-balanced occupations may reflect greater diversity of gender representation within these examples in training data.

For ambiguity categories, BLEU scores were highest for long unambiguous sentences (49.16\%), followed by unambiguous (40.51\%) and then ambiguous (37.71\%) sentences. However, these differences likely reflect a bias in BLEU rather than a true model effect. As shown in Table \ref{apx:tab:avg_word_count}, sentences in the long unambiguous category were significantly longer (48.6 words on average) than those in unambiguous (22.2 words) or ambiguous (15.8 words) categories. BLEU score is highly sensitive to sentence length: errors in longer sentences exert less impact on the score than errors in shorter ones. In contrast, COMET results indicate higher scores for ambiguous sentences. This difference may stem from COMET’s robustness to lexical variations, particularly when predicting gendered forms of professions (e.g., '\textit{chirurgien}' vs. '\textit{chirurgienne}'), which BLEU treats as distinct tokens.

The BLEU score has consistently exhibited larger gaps both between genders, and across variables. We ascribe this to to its sensitivity to surface-level changes, such as the exact form of an occupation. COMET, being less affected by such lexical variations, provides a more nuanced measure of translation quality. However, COMET’s robustness to gendered differences can sometimes obscure disparities in model performance, making it less sensitive to errors related to gender agreement.

\begin{table*}[ht]
\centering
\begin{tabular}{lcccc}
\toprule
\textbf{Models} & \textbf{Baseline} & \textbf{Moral Prompting} & \textbf{Linguistic Prompting} & \textbf{Moral and Linguistic Prompting} \\
\midrule
Gemma2-2B       & 5   & 9   & 29   & 49  \\
Mistral-7B       & 0   & 1   & 1    & 2   \\
Llama3.1-8B       & 1   & 4   & 10   & 42  \\
Llama3.3-70B   & 0   & 19  & 12   & 86  \\
\bottomrule
\end{tabular}
\caption{Number of gender-inclusive french translated sentences detected using gender-inclusive indicators in different models in different promptings out of 806 real gender-inclusive sentences in FairTranslate.}
\label{tab:inclusive_sentences}
\end{table*}

\section{Specific Analysis of Occupational Terms}\label{sec:occ_terms}

\begin{figure}[ht]
    \centering
    \subfigure{
        \includegraphics[width=0.22\textwidth]{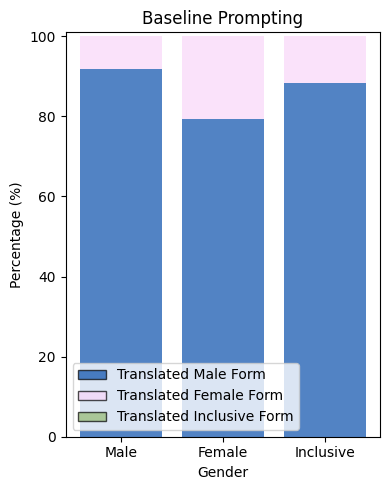}
    }
    \subfigure{
        \includegraphics[width=0.22\textwidth]{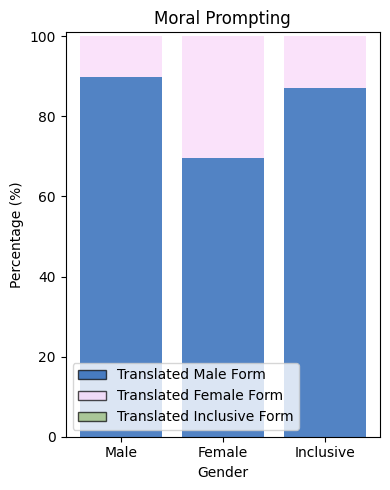}
    }
    \subfigure{
        \includegraphics[width=0.22\textwidth]{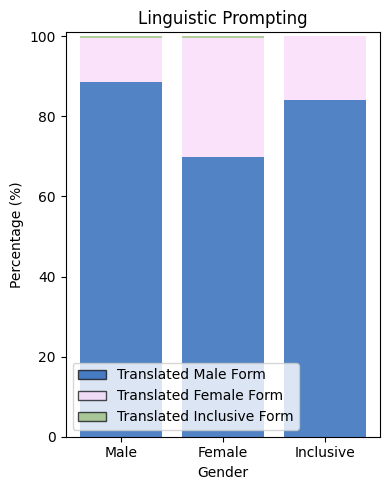}
    }
    \subfigure{
        \includegraphics[width=0.22\textwidth]{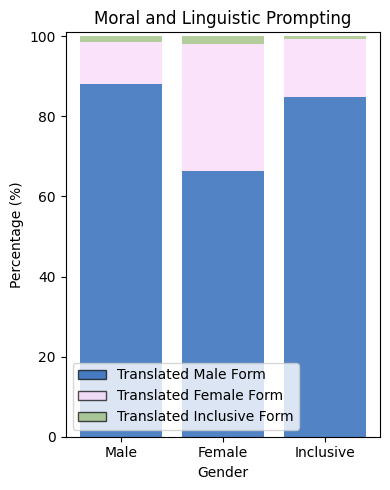}
    }
    \caption{Comparison of the Labelled Gender Based on the Form of the Translated Occupation by Gemma2-2B for Different Promptings.}
    \label{fig:comparison_gender_gemma2-2B}
\end{figure}

Accurately translating gendered forms of occupations in French addresses two key challenges: ensuring feminine forms are used for traditionally male-dominated occupations and incorporating modern inclusive forms to represent individuals beyond the binary gender spectrum. Both are critical for fairness and inclusivity in machine translation.

This experiment evaluates the Gemma2-2B model (other models are analyzed in appendix \ref{apx:sec:occ_gender}) in translating occupations from English to French, focusing on how gender is represented in the translated forms (as explained in Section \ref{sec:background}, French offers three gender grammatical forms). The goal is to assess whether the grammatical gender of the translated occupation aligns with the actual gender associated with the original sentence.

The dataset is filtered to retain only sentences where the correct occupation is translated by the model. For example, if the English sentence contains \textit{'nurse'} and the model translates it as \textit{'chirurgien'/'chirurgienne'/'chirurgien.ne'} (surgeon) instead of \textit{'infirmier'/'infirmière'/'infirmier.ère'} (nurse), the sentence is excluded.
The grammatical gender of the translated occupation is automatically identified using a predefined mapping, and it is compared to the original sentence's gender label. For instance, if the original sentence referred to a female nurse, the expected gender is feminine, and we check whether the translation uses the form \textit{'infirmière'}. This comparison allows us to measure how often the translation respects the intended gender.

Figure \ref{fig:comparison_gender_gemma2-2B} highlights a significant bias: the model overwhelmingly defaults to the masculine form, regardless of whether the original subject is male, female, or inclusive. Inclusive forms (e.g., infirmier.ière) are almost entirely absent, reflecting the model's inability to adopt recent linguistic innovations for inclusive representation. Attempts to improve inclusivity, such as "Moral Prompting" or "Linguistic Prompting," fail to resolve the issue and can even exacerbate it in some cases (see Appendix \ref{apx:sec:occ_gender}).


\begin{figure*}
    \centering
    \includegraphics[width=1\linewidth]{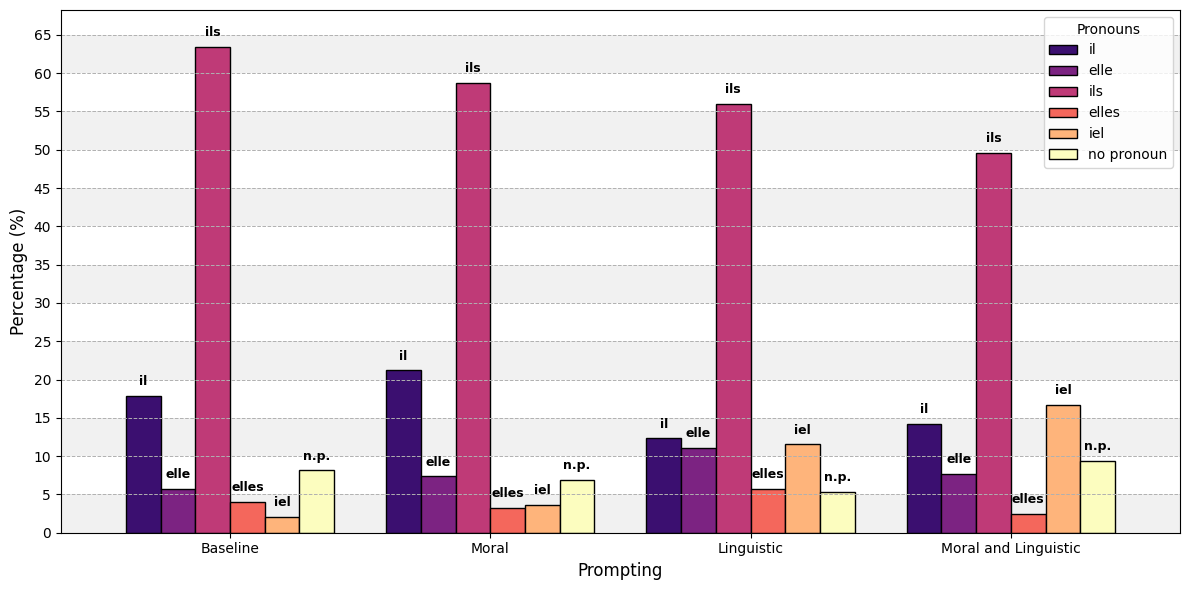}
    \caption{ Distribution of French Pronouns in the Translation of the Inclusive \textit{They} with Gemma2.
The figure is based on 246 English sentences containing the pronoun '\textit{they}' labeled as 'inclusive' and shows the translation results under the four promptings. 
The predominance of '\textit{ils}' in the figure highlights a bias in the model, which struggles to fully capture the inclusive aspect of '\textit{they}'.}
\label{fig:theypronouns_gemma}
\end{figure*}

\section{Specific Analysis of Inclusive Gender}

In Section \ref{sec:translation_evaluation}, we showed that models perform significantly worse when translating inclusive gender forms compared to masculine and feminine forms across all configurations. Additionally, in Section \ref{sec:occ_terms}, we observed that occupational terms are almost never translated into inclusive forms in French. Here, we extend the analysis to all types of inclusive gender indicators.

\subsection{Inclusive Gender Indicators}

We first investigate whether the poor translation of occupational terms into inclusive forms reflects a broader difficulty in translating inclusive gender indicators overall. Table \ref{tab:inclusive_sentences} presents the frequency of inclusive gender indicators appearing in French translations for various models and prompting strategies.

Inclusive gender indicators include terms such as ["Iel", "iel", "Lea", "lea", "Un.e", "un.e", "Ce.tte", "ce.tte"], as well as inclusive occupational forms identified by endings like ["ien.ne", "ier.ère", "eur.euse", "eur.e", "eux.euse", "tre.esse", "te.esse", "eur.rice"]. These indicators are assessed across 806 sentences annotated as inclusive in the gender column of the dataset. Each of these sentences in the FairTranslate dataset includes at least one inclusive indicator or occupational form in its correct French translation.

However, the models rarely produce the expected inclusive indicators or forms. Across all configurations, the number of sentences containing correctly translated inclusive forms ranges between 0 and 86 out of 806, which represents less than 11\% of sentences in the best case.

Although this performance is insufficient, we observe a trend: the use of moral or linguistic prompting improves the translation of inclusive forms compared to the baseline. Furthermore, combining both types of prompting provides the best results for all models tested.

\subsection{Inclusive They Translation Analysis}

To understand the errors models make when translating inclusive gender, we conduct a focused analysis of the singular '\textit{they}'. If models fail to translate inclusive indicators, what do they produce instead? 
For this analysis, we focus on the singular '\textit{they}' and examine 246 examples where it appears explicitly in the English source. Cases involving other inclusive markers (e.g., '\textit{their}') or ambiguities visible only in the French translation are excluded.

The singular '\textit{they}' can be translated into French in various ways by a model, but not all of which are correct or appropriate:
\begin{itemize}
    \item \textbf{Plural forms:} such as '\textit{ils}' (masculine plural), '\textit{elles}' (feminine plural) or '\textit{iels}' (inclusive plural). This suggests the model has not recognized the singular usage of '\textit{they}'.
    \item \textbf{Singular gendered forms:} such as '\textit{il}' (masculine singular) or '\textit{elle}' (feminine singular). This suggests the model understands the singular usage of '\textit{they}' but does not account for inclusive gender practices in French.
    \item \textbf{Inclusive forms:} such as '\textit{iel}', which reflects correct alignment with recent inclusive practices in French.
    \item \textbf{Pronoun omission:} a valid strategy in French for achieving inclusivity without explicitly using a pronoun.
\end{itemize}

Figure \ref{fig:theypronouns_gemma} shows that Gemma2-2B predominantly translates singular '\textit{they}' as a plurial. Prompting strategies improve the use of inclusive forms (e.g., '\textit{iel}' or omission) but remain insufficient, with plural translations still occurring in nearly 50\% of cases. Appendix \ref{apx:sec:inclusivethey} provides similar results for other models, often showing even worse performance.


These results reveal that the issue extends beyond models failing to adopt the relatively new inclusive practices in French. The problem is deeper: models often fail to correctly interpret inclusive usages in English, such as the singular '\textit{they}', which they misinterpret as plural\footnotetext{Note that in some cases, the use of the plural may be justified, as the singular '\textit{they}' of some English examples can lead to a double interpretation. However, this is not a majority of the examples in the dataset, so there is a real bias towards over-use of this translation.}. This fundamental misunderstanding highlights that the challenge is not merely one of cultural or linguistic adaptation to French but also reflects a lack of comprehension of long-established inclusive constructs in English. Addressing this requires substantial improvements in how models process and represent nuanced linguistic phenomena.

\section{Conclusion}
In this work, we introduced FairTranslate, a novel English-French translation dataset annotated by experts with extensive metadata, enabling detailed analysis of non-binary gender biases in translation. This dataset is a valuable resource for evaluating machine translation systems, particularly in handling inclusive gender forms. Additionally, we evaluated four LLMs using varied prompting strategies which has offered novel insights into how these models process gender inclusivity in translation tasks.

Our findings reveal consistent shortcomings in the translation of inclusive gender forms. Across all configurations, models performed significantly worse on inclusive gender translations compared to masculine and feminine forms. Moreover, French inclusive gender indicators and forms were almost never used, even when prompting was applied. This underscored the models' inability to integrate relatively recent linguistic practices in French.

Crucially, our analysis demonstrates that the inclusive language challenges extend beyond challenges which are specific to French. The poor translation of inclusive gender forms is rooted in a more fundamental problem: models fail to adequately understand inclusive constructs in English, such as the singular '\textit{they}'. The misinterpretation of these constructs as plural contributes to their inadequate French translations. This finding highlights the need for LLMs to better capture the nuances of inclusive language, both in English as well as in other languages such as French.

By publicly sharing our dataset and analysis, we aim to encourage the development of equitable translation systems that are not only linguistically competent but also inclusive, addressing biases that disproportionately affect underrepresented linguistic forms.

\begin{acks}
The authors thank all the people and industrial partners involved in the FOR and DEEL projects. This work has benefited from the support of the FOR\footnote{\url{https://www.irt-saintexupery.com/fr/for-program/}} and DEEL\footnote{\url{https://www.deel.ai/}} projects, with fundings from the Agence Nationale de la Recherche, and which is part of the ANITI AI cluster.


A special thank you goes to Daniel Anadria for his help and insightful suggestions throughout the writing process. His careful proofreading and constructive feedback greatly enhanced the clarity and overall quality of this manuscript.
\end{acks}


\newpage
\bibliographystyle{ACM-Reference-Format}
\bibliography{references}

\appendix

\section{Average Number of Words per Example in FairTranslate}
Table \ref{apx:tab:avg_word_count} shows the average number of words (in both English and French) per example in the FairTranslate dataset, categorized by gender ambiguity type. Sentences labeled as "ambiguous" are the shortest, with an average of 15.7 words in French and 13.6 in English, as they contain no gender information in English. These are followed by sentences labeled as "unambiguous," which average 22.2 words in French and 19.1 in English. The longest sentences are those labeled as "long unambiguous," designed to include a gender indicator in English placed as far as possible from the profession (which is gendered in the French translation) to study long-distance coreference resolution. These sentences average 48.6 words in French and 40.1 in English. It is worth noting that French sentences are generally longer than their English counterparts due to structural differences between the two languages.

\begin{table}[h!]
\centering
\begin{tabular}{lcc}
\hline
\textbf{Ambiguity}       & \textbf{French} & \textbf{English} \\\hline
Ambiguous                & 15.775          & 13.564          \\
Unambiguous              & 22.153          & 19.102          \\
Long Unambiguous         & 48.599          & 40.104          \\\hline
\textbf{\#Avg Word Count} & \textbf{28.842} & \textbf{24.257} \\\hline
\end{tabular}
\caption{Average Word Count for French and English Sentences by Ambiguity Category in FairTranslate Dataset.}
\label{apx:tab:avg_word_count}
\end{table}

\section{ANOVA tests}
Table \ref{apx:tab:anovatest} presents the p-values from all ANOVA tests conducted to verify whether the differences in scores (for BLEU and COMET) across genders are statistically significant. All p-values are well below the 0.05 threshold, with every value being smaller than $10^{-10}$.

\begin{table*}[ht]
\centering
\begin{tabular}{lcccccccc}
\toprule
\multirow{2}{*}{Prompting} & \multicolumn{2}{c}{Gemma2-2B} & \multicolumn{2}{c}{Mistral-7B} & \multicolumn{2}{c}{Llama3.1-8B} & \multicolumn{2}{c}{Llama3.3-70B} \\
\cmidrule(lr){2-3} \cmidrule(lr){4-5} \cmidrule(lr){6-7} \cmidrule(lr){8-9}
 & BLEU & COMET & BLEU & COMET & BLEU & COMET & BLEU & COMET \\
\midrule
Baseline & $8.9\times 10^{-20}$ & $2.8\times 10^{-46}$ & $7.6\times 10^{-16}$ & $1.6\times 10^{-40}$ & $1.2\times 10^{-19}$ & $1.5\times 10^{-67}$ & $3.2\times 10^{-25}$ & $2.6\times 10^{-91}$ \\
Moral & $1.9\times 10^{-15}$ & $8.6\times 10^{-45}$ & $3.9\times 10^{-15}$ & $2.6\times 10^{-46}$ & $7.1\times 10^{-15}$ & $6.1\times 10^{-59}$ & $4.1\times 10^{-11}$ & $5.6\times 10^{-18}$ \\
Linguistic & $1.6\times 10^{-11}$ & $1.4\times 10^{-34}$ & $1.8\times 10^{-14}$ & $1.1\times 10^{-35}$ & $6.7\times 10^{-16}$ & $1.5\times 10^{-48}$ & $6.1\times 10^{-23}$ & $3.6\times 10^{-78}$ \\
Moral \& Ling. & $9.6\times 10^{-15}$ & $6.5\times 10^{-38}$ & $5.4\times 10^{-14}$ & $5.6\times 10^{-36}$ & $8.4\times 10^{-13}$ & $3.1\times 10^{-39}$ & $2.3\times 10^{-20}$ & $1.8\times 10^{-74}$ \\
\bottomrule
\end{tabular}
\caption{P-values from ANOVA tests across gender categories (female, male, inclusive) for BLEU and COMET scores, on different models and prompting configurations.}
\label{apx:tab:anovatest}
\end{table*}

\section{Effects of Prompting for COMET scores}\label{apx:sec:prompting_comet}

\begin{table*}[h!]
\centering
\begin{tabular}{lcccccccccccc}
\toprule
\textbf{Prompting} & \multicolumn{3}{c}{\textbf{Gemma2-2B}} & \multicolumn{3}{c}{\textbf{Mistral-7B}} & \multicolumn{3}{c}{\textbf{Llama3.1-8B}} & \multicolumn{3}{c}{\textbf{Llama3.3-70B}} \\
\cmidrule(lr){2-4} \cmidrule(lr){5-7} \cmidrule(lr){8-10} \cmidrule(lr){11-13}
 & Female & Male & Inclusive 
 & Fem. & Male & Incl. & Fem. & Male & Incl.& Fem. & Male & Incl. \\
\midrule
Baseline & 87.86 & \textbf{89.18} & \textbf{85.42} & \textbf{87.16} & 88.25 & \textbf{84.42} & \textbf{88.88} & \textbf{89.85} & 85.97 & 90.12 & \textbf{91.35} & 87.16 \\
Moral & 87.80 & 89.16 & \textbf{85.42} & 86.82 & \textbf{88.41} & 84.30 & 88.45 & 89.83 & 86.15 & 87.95 & 89.20 & 86.52 \\
Linguistic & 87.59 & 88.81 & 85.39 & 86.18 & 87.92 & 84.03 & 88.86 & 89.39 & 86.16 & \textbf{90.19} & 91.26 & 87.59 \\
Moral \& Ling. & \textbf{87.96} & 88.78 & 85.32 & 86.43 & 87.93 & 84.17 & 88.68 & 89.20 & \textbf{86.23} & 89.85 & 91.26 & \textbf{87.63}  \\
\bottomrule
\end{tabular}
\caption{COMET scores for different promptings and genders. The bold font means the Prompting performs best for the gender and the model.}
\label{apx:tab:prompting_comet_scores}
\end{table*}

Table \ref{apx:tab:prompting_comet_scores} evaluates the effect of four prompting strategies—baseline, moral, linguistic, and moral+linguistic—on translation performance, focusing on COMET scores across gender categories.

\begin{itemize}
    \item Male Gender: Performance generally decreases as prompting strategies incorporate more information (only one exception: Mistral-7B, which shows a slight improvement under the moral prompting).
    \item Female Gender: Performance also generally decreases (two exceptions: Gemma2-2B with Moral\&Linguistic and Llama3.3-70B with Linguistic).
    \item Inclusive Gender: Performance decreases for Gemma and Mistral, but increases for Llama3.1 and Llama3.3 with linguistic and combined prompts (except under the Moral prompting for Llama3.3).
\end{itemize}

While overall performance tends to decline across all genders (except for inclusive gender on 2 models) with increased prompting, the decline is most pronounced for the male gender, reducing disparities between genders. However, these reductions remain insufficient for equitable translation performance.

\section{Analysis of Metadata Variables for All Models}\label{apx:sec:metavar}

Tables \ref{apx:tab:stereotype_scores} and \ref{apx:tab:ambiguity_scores} report the performance of all models across gender categories with respect to the stereotype and ambiguity variables, respectively. The results from these tables align closely with those observed for Gemma2-2B in Table \ref{tab:stereo_ambig_scores_gemma2}. This consistency confirms that the analysis presented in Section \ref{sec:metavar} generalizes well across all models in our experimental setup.

\begin{table*}[h!]
    \centering
    \begin{tabular}{@{}l@{\;}lcccccccccc@{}}
        \toprule
        \multirow{2}{*}{\parbox{1.7cm}{\centering \makecell{Stereotype}}} & \multirow{1.6}{*}{Gender} & \multicolumn{2}{c}{Gemma2-2B} & \multicolumn{2}{c}{Mistral-7B} & \multicolumn{2}{c}{Llama3.1-8B} & \multicolumn{2}{c}{Llama3.3-70B} & \multicolumn{2}{c}{Mean} \\
        \cmidrule(lr){3-4} \cmidrule(lr){5-6} \cmidrule(lr){7-8} \cmidrule(lr){9-10} \cmidrule(lr){11-12}
         &  & BLEU & COMET & BLEU & COMET & BLEU & COMET & BLEU & COMET & BLEU & COMET \\
        \midrule
        \multirow{4}{*}{\makecell{Female-\\stereotyped}} & Female & 40.78 & 87.84 & 37.33 & 86.79 & \textbf{43.15} & \textbf{89.06} & 48.79 & 90.04 & 42.51 & 88.43 \\
         & Male & \textbf{42.35} & \textbf{88.33} & \textbf{38.91} & \textbf{86.89} & 42.64 & 88.88 & \textbf{50.54} & \textbf{90.17} & \textbf{43.61} & \textbf{88.57} \\
         & Inclusive & 34.88 & 84.60 & 32.56 & 82.95 & 35.45 & 85.11 & 42.18 & 85.84 & 36.27 & 84.63 \\
         \cmidrule(lr){2-12}
         & \textit{Mean} & \textit{39.34} & \textit{86.92} & \textit{36.27} & \textit{85.54} & \textit{40.41} & \textit{87.68} & \textit{47.17} & \textit{88.68} & \textit{40.80} & \textit{87.20} \\
        \midrule
        \multirow{4}{*}{\makecell{Male-\\stereotyped}} & Female & 41.33 & 87.13 & 38.46 & 86.70 & 42.70 & 88.15 & 52.21 & 89.92 & 43.68 & 87.98 \\
         & Male & \textbf{44.69} & \textbf{88.76} & \textbf{41.20} & \textbf{88.45} & \textbf{45.71} & \textbf{89.76} & \textbf{57.52} & \textbf{91.85} & \textbf{47.28} & \textbf{89.70} \\
         & Inclusive & 36.45 & 85.42 & 33.73 & 85.21 & 37.91 & 86.43 & 46.64 & 88.05 & 38.68 & 86.28 \\
         \cmidrule(lr){2-12}
         & \textit{Mean} & \textit{40.82} & \textit{87.10} & \textit{37.80} & \textit{86.79} & \textit{42.11} & \textit{88.11} & \textit{52.12} & \textit{89.94} & \textit{43.21} & \textit{88.49} \\
        \midrule
        \multirow{4}{*}{\makecell{Gender-\\balanced}} & Female & 45.29 & 88.72 & 41.34 & 88.12 & 45.56 & 89.48 & 53.63 & 90.44 & 46.45 & 89.19 \\
         & Male & \textbf{49.63} & \textbf{90.63} & \textbf{44.35} & \textbf{89.58} & \textbf{50.80} & \textbf{91.08} & \textbf{60.15} & \textbf{92.15} & \textbf{51.23} & \textbf{90.86} \\
         & Inclusive & 38.91 & 86.36 & 35.20 & 85.23 & 39.48 & 86.45 & 47.26 & 87.68 & 40.21 & 86.43 \\
         \cmidrule(lr){2-12}
         & \textit{Mean} & \textbf{\textit{44.61}} & \textbf{\textit{88.57}} & \textbf{\textit{40.30}} & \textbf{\textit{87.64}} & \textbf{\textit{45.28}} & \textbf{\textit{89.00}} & \textbf{\textit{53.01}} & \textbf{\textit{90.09}} & \textbf{\textit{45.30}} & \textbf{\textit{88.82}} \\
        \bottomrule
    \end{tabular}
    \caption{BLUE and COMET scores for each model and each combination of stereotype and gender on Baseline Prompting, with averages per model and per stereotype. The bold font means the gender performs best for the model. Bold italicized font means that the class in stereotype is the best-performing class for the model.}
    \label{apx:tab:stereotype_scores}
\end{table*}

\begin{table*}[h!]
    \centering
    \begin{tabular}{@{}l@{\;}lcccccccccc@{}}
        \toprule
        \multirow{2}{*}{\parbox{1.5cm}{\centering \makecell{Ambiguity}}} & \multirow{2}{*}{Gender} & \multicolumn{2}{c}{Gemma2-2B} & \multicolumn{2}{c}{Mistral-7B} & \multicolumn{2}{c}{Llama3.1-8B} & \multicolumn{2}{c}{Llama3.3-70B} & \multicolumn{2}{c}{Mean} \\
        \cmidrule(lr){3-4} \cmidrule(lr){5-6} \cmidrule(lr){7-8} \cmidrule(lr){9-10} \cmidrule(lr){11-12}
         &  & BLEU & COMET & BLEU & COMET & BLEU & COMET & BLEU & COMET & BLEU & COMET \\
        \midrule
        \multirow{4}{*}{Ambiguous} & Female & 36.65 & 88.25 & 34.09 & 87.69 & 38.69 & 89.08 & 44.45 & 90.40 & 38.47 & 88.85 \\
         & Male & \textbf{41.32} & \textbf{89.36} & \textbf{38.04} & \textbf{88.85} & \textbf{42.98} & \textbf{90.47} & \textbf{50.69} & \textbf{91.75} & \textbf{43.26} & \textbf{90.11} \\
         & Inclusive & 35.17 & 85.78 & 33.65 & 85.18 & 36.07 & 86.77 & 44.32 & 87.75 & 37.30 & 86.37 \\
         \cmidrule(lr){2-12}
         & \textit{Mean} & \textit{37.71} & \textbf{\textit{87.80}} & \textit{35.26} & \textbf{\textit{87.24}} & \textit{39.25} & \textbf{\textit{88.77}} & \textit{46.49} & \textbf{\textit{89.97}} & \textit{39.68} & \textit{\textbf{88.44}} \\
        \midrule
        \multirow{4}{*}{\makecell{Long\\Unambiguous}} & Female & 50.35 & 87.66 & 46.31 & 86.70 & 51.02 & 88.46 & 60.26 & 89.28 & 51.99 & 88.02 \\
         & Male & \textbf{51.66} & \textbf{88.77} & \textbf{47.76} & \textbf{87.33} & \textbf{51.12} & \textbf{88.74} & \textbf{62.15} & \textbf{90.21} & \textbf{53.17} & \textbf{88.76} \\
         & Inclusive & 45.48 & 86.67 & 42.79 & 85.80 & 46.53 & 87.10 & 55.08 & 88.00 & 47.47 & 86.89 \\
         \cmidrule(lr){2-12}
         & \textit{Mean} & \textbf{\textit{49.16}} & \textit{87.69} & \textbf{\textit{45.62}} & \textit{86.61} & \textbf{\textit{49.56}} & \textit{88.10} & \textbf{\textit{59.16}} & \textit{89.16} & \textit{\textbf{50.87}} & \textit{87.89} \\
        \midrule
        \multirow{4}{*}{Unambiguous} & Female & 43.17 & 87.60 & 39.30 & 86.92 & 44.33 & 88.92 & 53.06 & 90.35 & 44.97 & 88.45 \\
         & Male & \textbf{45.61} & \textbf{89.25} & \textbf{40.79} & \textbf{88.19} & \textbf{46.39} & \textbf{89.91} & \textbf{57.22} & \textbf{91.64} & \textbf{47.50} & \textbf{89.75} \\
         & Inclusive & 32.76 & 84.30 & 28.38 & 82.83 & 33.51 & 84.49 & 40.24 & 86.06 & 33.72 & 84.42 \\
         \cmidrule(lr){2-12}
         & \textit{Mean} & \textit{40.51} & \textit{87.05} & \textit{36.16} & \textit{85.98} & \textit{41.41} & \textit{87.77} & \textit{50.17} & \textit{89.35} & \textit{42.06} & \textit{87.54} \\
        \bottomrule
    \end{tabular}
    \caption{Mean BLEU and COMET scores (in percentage) for each model and each combination of ambiguity and gender. The bold font means the gender performs best for the model. Bold italicized font means that the class in ambiguity is the best-performing class for the model.}
    \label{apx:tab:ambiguity_scores}
\end{table*}

\newpage
\section{Specific Analysis of Occupational Terms for All Models}\label{apx:sec:occ_gender}

\begin{figure*}[h!]
    \centering
    \subfigure{
        \includegraphics[width=0.22\textwidth]{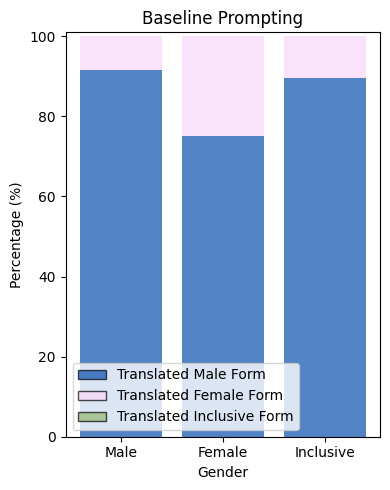}
    }
    \subfigure{
        \includegraphics[width=0.22\textwidth]{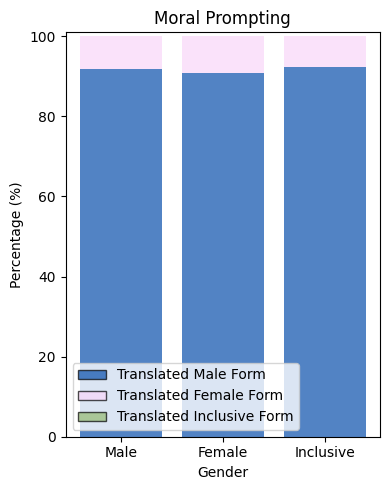}
    }
    \subfigure{
        \includegraphics[width=0.22\textwidth]{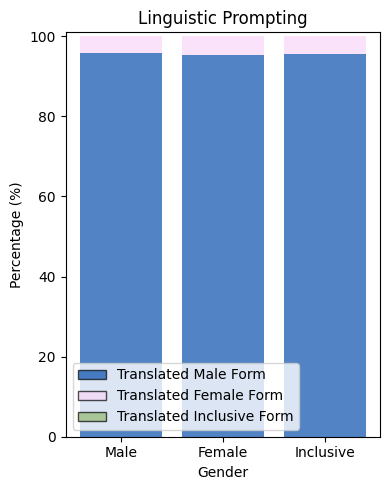}
    }
    \subfigure{
        \includegraphics[width=0.22\textwidth]{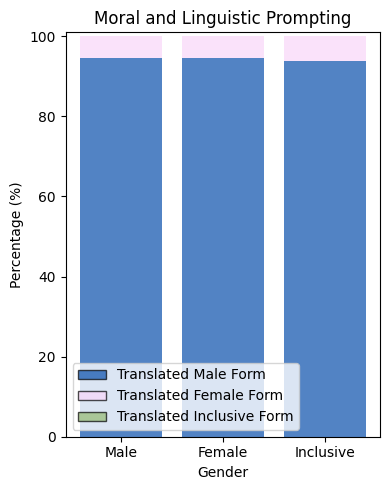}
    }
    \caption{Comparison of the Labelled Gender Based on the Form of the Translated Occupation by Mistral-7B for Different Promptings.}
    \label{apx:fig:comparison_gender_mistral-7B}
\end{figure*}

\begin{figure*}[h!]
    \centering
    \subfigure{
        \includegraphics[width=0.22\textwidth]{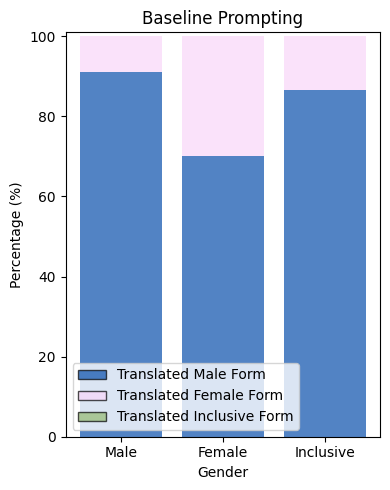}
    }
    \subfigure{
        \includegraphics[width=0.22\textwidth]{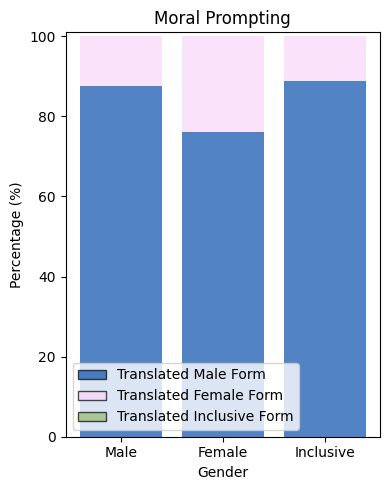}
    }
    \subfigure{
        \includegraphics[width=0.22\textwidth]{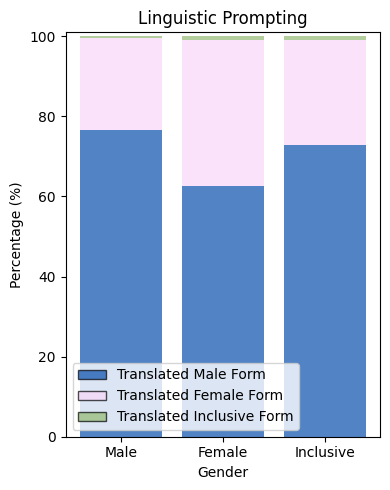}
    }
    \subfigure{
        \includegraphics[width=0.22\textwidth]{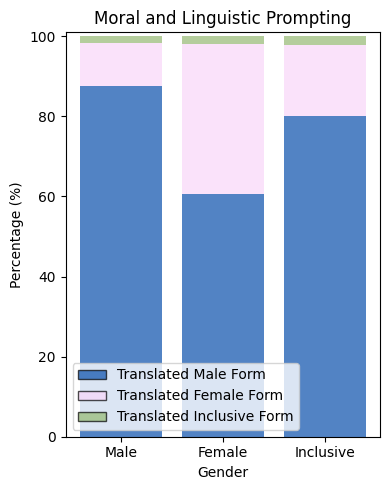}
    }
    \caption{Comparison of the Labelled Gender Based on the Form of the Translated Occupation by Llama3.1-8B for Different Promptings.}
    \label{apx:fig:comparison_gender_llama31-8B}
\end{figure*}

\begin{figure*}[h!]
    \centering
    \subfigure{
        \includegraphics[width=0.22\textwidth]{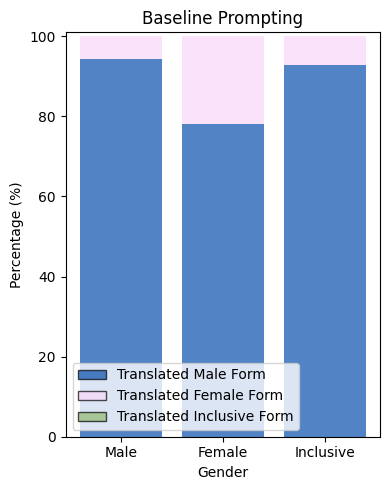}
    }
    \subfigure{
        \includegraphics[width=0.22\textwidth]{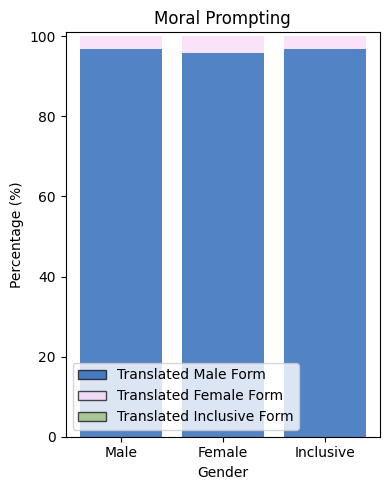}
    }
    \subfigure{
        \includegraphics[width=0.22\textwidth]{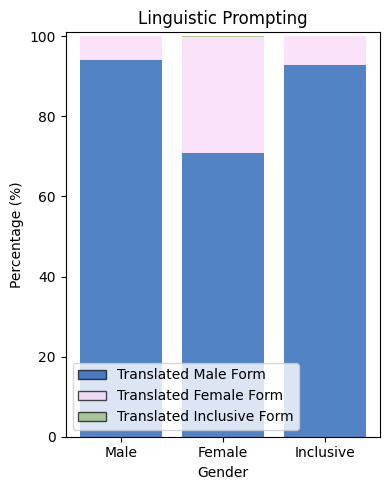}
    }
    \subfigure{
        \includegraphics[width=0.22\textwidth]{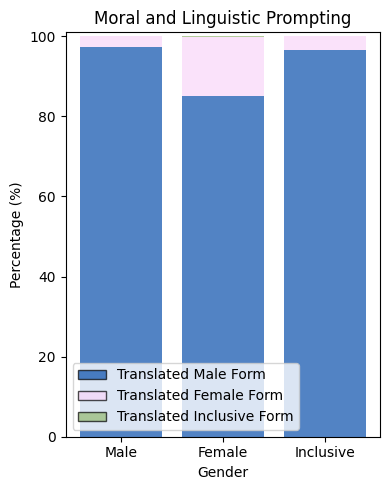}
    }
    \caption{Comparison of the Labelled Gender Based on the Form of the Translated Occupation by Llama3.3-70B for Different Promptings.}
    \label{apx:fig:comparison_gender_llama33-70B}
\end{figure*}

Figures \ref{apx:fig:comparison_gender_mistral-7B}, \ref{apx:fig:comparison_gender_llama31-8B}, and \ref{apx:fig:comparison_gender_llama33-70B} illustrate the percentage of occupations translated into their masculine (blue), feminine (pink), and inclusive (green) forms for sentences labeled with masculine, feminine, and inclusive occupational forms in the source language. The analysis is presented for three models: Mistral-7B, Llama3.1-8B, and Llama3.3-70B.

For Mistral-7B and Llama3.3-70B, prompting exacerbates the issue. Not only do these models fail to produce any inclusive translations of occupations, but the percentage of occupations translated into the feminine form also decreases.

In contrast, Llama3.1-8B behaves similarly to Gemma2-2B, showing a slight improvement in the percentage of occupations in the feminine form being correctly translated and a marginal introduction of inclusive forms. However, these results remain far from satisfactory, highlighting significant gaps in gender representation in the translations.

\newpage

\section{Inclusive They Translation Analysis for All Models}\label{apx:sec:inclusivethey}

\begin{figure*}
    \centering
    \includegraphics[width=1\linewidth]{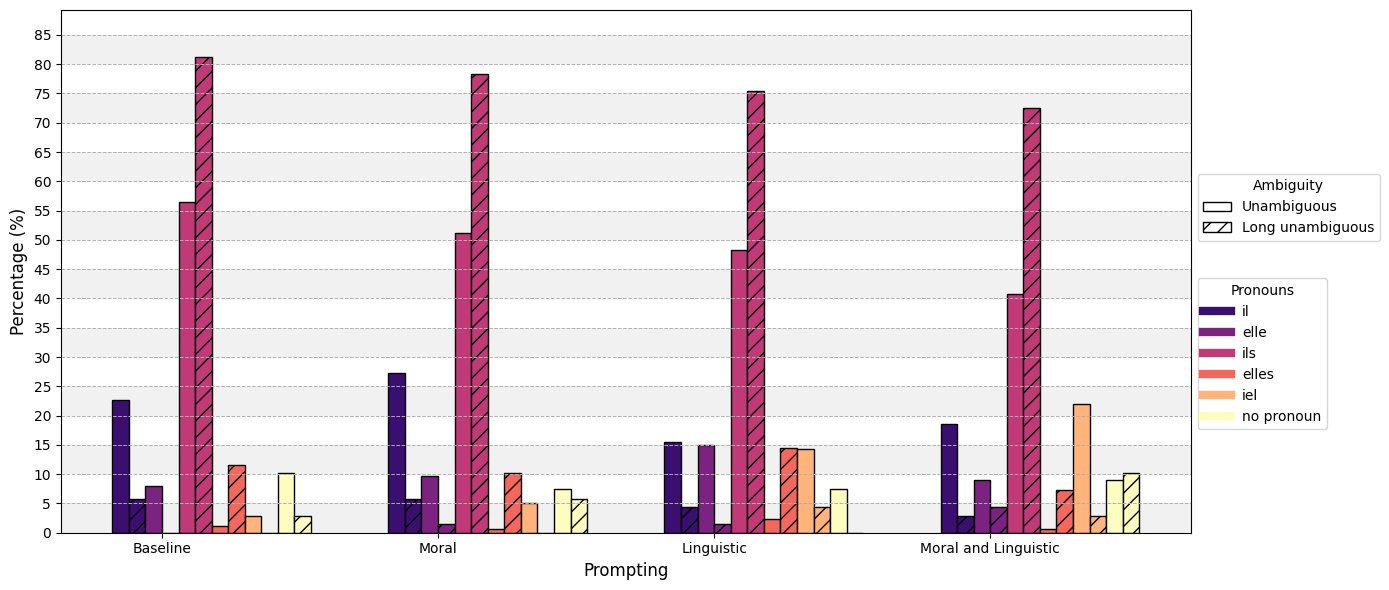}
    \caption{
\textbf{Distribution of French Pronouns in the Translation of the Inclusive \textit{They} with Gemma2-2B by ambiguity variable.}
The figure is based on 177 unambiguous and 69 long unambiguous English sentences containing the pronoun '\textit{they}' and shows the translation results under the four promptings. 
The appropriate translations for the inclusive '\textit{they}' are typically either '\textit{iel}' or constructing the sentence without a pronoun (\textit{no pronom}). The predominance of '\textit{ils}' in the figure highlights a bias in the model, which struggles to fully capture the inclusive aspect of '\textit{they}'.}
    \label{apx:fig:theypronouns_gemma_ambiguity}
\end{figure*}

\begin{figure*}
    \centering
    \includegraphics[width=1\linewidth]{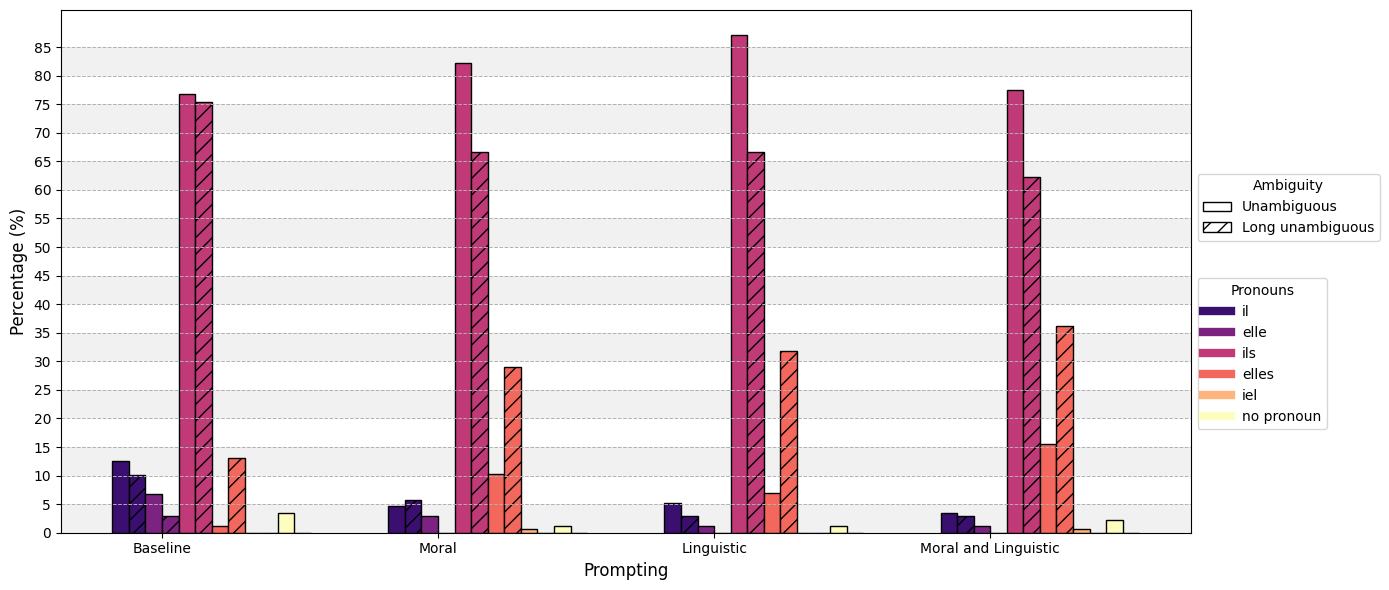}
    \caption{
\textbf{Distribution of French Pronouns in the Translation of the Inclusive \textit{They} with Mistral-7B by ambiguity variable.}
The figure is based on 177 unambiguous and 69 long unambiguous English sentences containing the pronoun '\textit{they}' and shows the translation results under the four promptings. 
The appropriate translations for the inclusive '\textit{they}' are typically either '\textit{iel}' or constructing the sentence without a pronoun (\textit{no pronom}). The predominance of '\textit{ils}' in the figure highlights a bias in the model, which struggles to fully capture the inclusive aspect of '\textit{they}'.}
    \label{apx:fig:theypronouns_mistral_ambiguity}
\end{figure*}

\begin{figure*}
    \centering
    \includegraphics[width=1\linewidth]{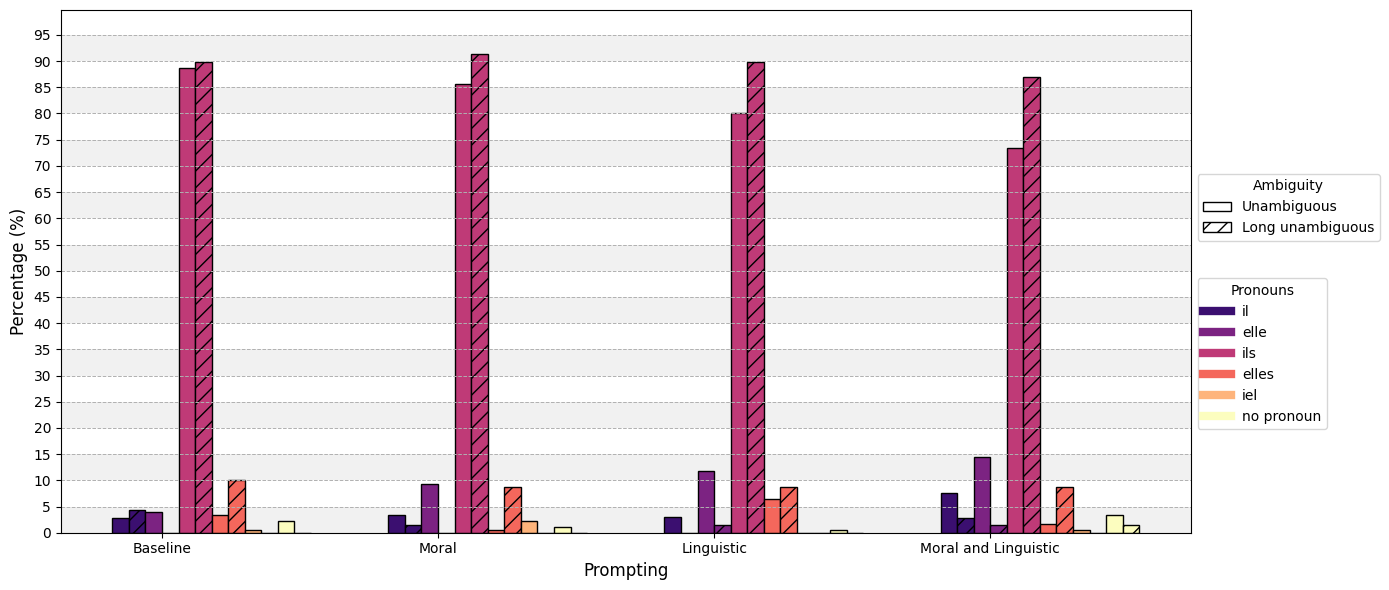}
    \caption{
\textbf{Distribution of French Pronouns in the Translation of the Inclusive \textit{They} with Llama3.1-8B by ambiguity variable.}
The figure is based on 177 unambiguous and 69 long unambiguous English sentences containing the pronoun '\textit{they}' and shows the translation results under the four promptings. 
The appropriate translations for the inclusive '\textit{they}' are typically either '\textit{iel}' or constructing the sentence without a pronoun (\textit{no pronom}). The predominance of '\textit{ils}' in the figure highlights a bias in the model, which struggles to fully capture the inclusive aspect of '\textit{they}'.}
    \label{apx:fig:theypronouns_llama31_ambiguity}
\end{figure*}

\begin{figure*}
    \centering
    \includegraphics[width=1\linewidth]{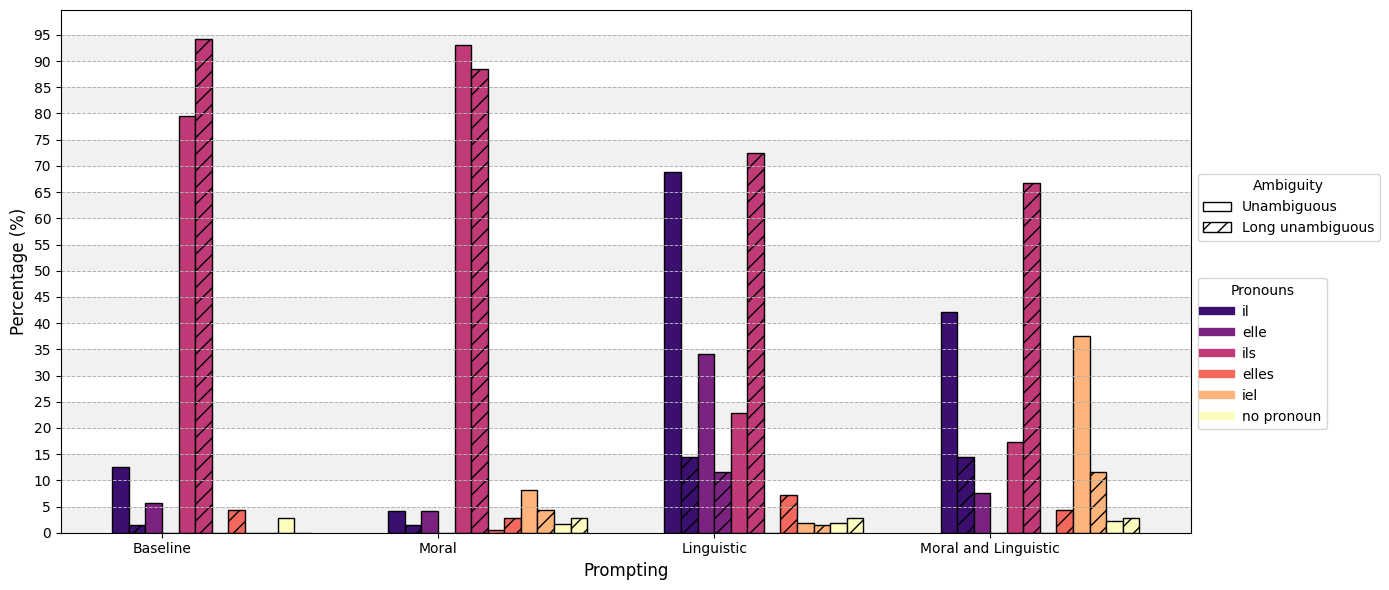}
    \caption{
\textbf{Distribution of French Pronouns in the Translation of the Inclusive \textit{They} with Llama3.3-70B by ambiguity variable.}
The figure is based on 177 unambiguous and 69 long unambiguous English sentences containing the pronoun '\textit{they}' and shows the translation results under the four promptings. 
The appropriate translations for the inclusive '\textit{they}' are typically either '\textit{iel}' or constructing the sentence without a pronoun (\textit{no pronom}). The predominance of '\textit{ils}' in the figure highlights a bias in the model, which struggles to fully capture the inclusive aspect of '\textit{they}'.}
    \label{apx:fig:theypronouns_llama33_ambiguity}
\end{figure*}

Figures \ref{apx:fig:theypronouns_gemma_ambiguity}, \ref{apx:fig:theypronouns_mistral_ambiguity}, \ref{apx:fig:theypronouns_llama31_ambiguity} and \ref{apx:fig:theypronouns_llama33_ambiguity} present the distribution of French pronouns in the translation of the singular '\textit{they}', filtered by the ambiguity variable. Only the "long unambiguous" and "unambiguous" cases are considered, as "ambiguous" cases are excluded since the singular '\textit{they}' cannot appear in sentences without any gender indicators in English. This analysis evaluates whether context length affects the accurate translation of singular '\textit{they}'.

For Gemma2-2B (Figure \ref{apx:fig:theypronouns_gemma_ambiguity}), the results show that in "long unambiguous" cases, where there is a long context between the occupation (singular) and '\textit{they}', the model translates '\textit{they}' as a plural pronoun over 80\% of the time. In contrast, when '\textit{they}' appears close to the singular occupation ("unambiguous"), the plural translation rate drops to 50\%, indicating that the model is better able to resolve the link in shorter contexts.

This pattern is much weaker for Mistral-7B (Figure \ref{apx:fig:theypronouns_mistral_ambiguity}) and Llama3.1-8B (Figure \ref{apx:fig:theypronouns_llama31_ambiguity}), which nearly always translate '\textit{they}' as a plural pronoun, regardless of context length or prompting strategy.

For Llama3.3-70B (Figure \ref{apx:fig:theypronouns_llama33_ambiguity}), the baseline and moral prompting strategies yield results similar to Mistral-7B and Llama3.1-8B. However, in "unambiguous" cases (short contexts), applying a linguistic prompting strategy causes the model to translate '\textit{they}' as a singular pronoun, albeit with gendered markers like '\textit{il}' or '\textit{elle}.' This represents a first step toward accurate translation. When both linguistic and moral context prompts are combined, the model more often successfully produces the correct translation of the inclusive singular, '\textit{iel}'. But the more complex case with a long context ("long unambiguous") is still just as bad.

\end{document}